\documentclass[11pt,a4paper]{article}
\usepackage[hyperref]{acl2021}

\usepackage{times}
\usepackage{graphicx}
\usepackage{latexsym}
\usepackage{amsmath}
\DeclareMathOperator*{\argmax}{arg\,max}

\usepackage{gb4e}
\noautomath
\usepackage{microtype}
\usepackage{subcaption}

\aclfinalcopy 
\def\aclpaperid{3} 

\newcommand{\quotes}[1]{``#1''}
\newcommand{\urlNewWindow}[1]{\href[pdfnewwindow=true]{#1}{\nolinkurl{#1}}}

\title{Let's be explicit about that: Distant supervision for implicit discourse relation classification via connective prediction}

\author{Murathan Kurfalı\\
  Department of Linguistics \\
  Stockholm University\\
  Stockholm, Sweden \\
\texttt{murathan.kurfali@ling.su.se } \\\And
Robert Östling\\
  Department of Linguistics \\
  Stockholm University\\
  Stockholm, Sweden \\
\texttt{robert@ling.su.se } 
}

\date{}

\begin{document}
\maketitle

\begin{abstract}

In implicit discourse relation classification, we want to predict the relation between adjacent sentences in the absence of any overt discourse connectives. This is challenging even for humans, leading to shortage of annotated data, a fact that makes the task even more difficult for supervised machine learning approaches. In the current study, we perform implicit discourse relation classification without relying on any labeled implicit relation. We sidestep the lack of data through explicitation of implicit relations to reduce the task to two sub-problems: language modeling and explicit discourse relation classification, a much easier problem. Our experimental results show that this method can even marginally outperform the state-of-the-art, in spite of being much simpler than alternative models of comparable performance. Moreover, we show that the achieved performance is robust across domains as suggested by the zero-shot experiments on a completely different domain. This indicates that recent advances in language modeling have made language models sufficiently good at capturing inter-sentence relations without the help of explicit discourse markers.

\end{abstract}

\section{Introduction}

Discourse relations describe the relationship between discourse units, e.g.\ clauses or sentences. These relations are either signalled explicitly with a discourse connective (e.g.\ \textit{because}, \textit{and}) or expressed implicitly and are inferred by sequential reading (Example \ref{ex:1} below). 

\begin{exe} \label{ex:1}
\item A figure above 50 indicates the economy is likely to expand. \textbf{[While]}	One below 50 indicates a contraction may be ahead. \textit{(Comparison - wsj\_0233)}
\end{exe}

The relations in the latter category are called \textit{implicit discourse relations} and they are of special significance because their lack of an explicit signal makes them challenging to annotate for even humans, suggested by the lower inter-annotator agreements on implicit relations \cite{zeyrek2017tdb,zikanova2019explicit}, let alone classify automatically. 

Resources for implicit discourse relations, therefore, are very limited. Even the Penn Discourse Tree Bank 2.0 (PDTB 2.0) \cite{prasad2008penn}, which is the most popular resource, includes merely 16K implicit discourse relations, all annotated on the same domain. Explicit discourse relations, on the other hand, are proven to be simple enough to be obtained both manually and automatically. Previous work shows that explicit relations in English have a low level of ambiguity, so the discourse relation can be classified with more than 94\% accuracy from the discourse connective alone \cite{pitler2009using}. This has inspired others to predict connectives for the implicit discourse relations and add them as additional features to existing supervised classifiers \cite{zhou2010predicting,xu2012connective}.

Our work takes this idea one step further by reducing the amount of supervision required. Instead of training a separate connective classifier, we generate a set of \textit{candidate explicit relations} that are obtained by inserting explicit discourse markers between sentences and score the resulting segments using a large pre-trained language model.\footnote{In the reminder of the paper, these candidate explicit relations are simply referred as \textit{candidates}.} The candidates are then classified with an accurate explicit discourse relation classifier, and the final implicit relation prediction can be obtained by either using the candidate with the highest-scoring connective, or marginalizing over the whole distribution of explicit connectives.

The main contributions of our papers are as follows:
\begin{itemize}
    \item We show that this simple approach is very effective and even marginally outperforms the current state-of-the-art method that does not use labeled implicit discourse relation data, even though that method uses a significantly more complex adversarial domain adaptation model \cite{huang2019unsupervised}.
    \item To the best of our knowledge, this is the first study to go beyond the default four-way classification under the low-resource scenario assumption where no labeled implicit discourse relation is available. We show that the proposed pipeline maintains its performance (relative to the baselines) in a more challenging 11-way classification as well as across domains (i.e., biomedical texts \cite{prasad2011biomedical}). 
    \item We offer explicitation of implicit discourse relations as a probing task to evaluate language models. Despite their relevancy, discourse relations are mostly overlooked in the assessments of language models' understanding of context. As a secondary aim, we investigate a wide range of pre-trained language models' understanding of inter-sentential relations.
\end{itemize}

We hope that the proposed pipeline will be another step in overcoming the data-bottleneck problem in discourse studies.

\section{Background}

\subsection{Implicit Discourse Relations}

PDTB 2.0 adopts a lexicalized approach where each relation consists of a discourse connective (e.g. \quotes{but}, \quotes{and}) which acts as a predicate taking two arguments. For each relation, annotators were asked to annotate the connective, the two text spans that hold the relation and the sense it conveys based on the PDTB sense hierarchy \cite{prasad2008penn}. The text span which is syntactically bound to the connective is called the second argument (arg2) whereas the other is the first argument (arg1). 
"Additionally, implicit relations are annotated with
that explicit connective which according to judgements
best expresses the sense of the relation."

However, in certain cases, a relation holds between the adjacent sentences despite the lack of an overt connective (see Example \ref{ex:1}). PDTB 2.0 recognizes such relations as \textit{implicit discourse relations}. Additionally, implicit relations are annotated with an explicit connective which best expresses the sense of the relation is according to annotators. The connective inserted by the annotators is termed as \quotes{\textit{implicit connective}} (e.g. \quotes{while} in Example \ref{ex:1}). Unlike explicit relations where there is an explicit textual cue (the connective), implicit relations can only be inferred which makes them more challenging to spot and annotate.

\subsection{Related Work}

The research on implicit discourse relation classification is overwhelmingly supervised \cite{pitler2009automatic,rutherford2015improving,lan2017multi,nie2019dissent,kim2020implicit}. Although unsupervised methods were present in the earliest attempts \cite{marcu2002unsupervised}, they haven't received serious attention and much research concentrated on increasing the available supervision to deal with the data; most prominently, either by automatically generating artificial data \cite{sporleder2008using,braud2014combining,rutherford2015improving,wu2016bilingually,shi2017using} or through introducing auxiliary but similar tasks to the training routine to leverage additional information \cite{zhou2010predicting,xu2012connective,liu2016implicit,lan2017multi,qin2017adversarial,shi2019learning,nie2019dissent}. \newcite{zhou2010predicting} and \newcite{xu2012connective} constitute the earliest examples where the classification of implicit relations are assisted via connective prediction. Both studies employ language models to predict suitable connectives for implicit relations which are, then, either used as additional features or classified directly.

\newcite{ji2015closing} is one of the few recent distantly supervised\footnote{Previous work uses the term \emph{unsupervised} (domain adaptation). Although we use the same amount of supervision with earlier work (no labeled implicit relation are utilized), we believe \emph{distant supervision} describes the method better.} studies which tackle implicit relation classification as a domain adaptation problem where the labeled explicit relations are regarded as the source domain and the unlabeled implicit relations as the target. \newcite{huang2019unsupervised} improves upon \newcite{ji2015closing} by employing adversarial domain adaption with a novel reconstruction component.

\subsection{Pre-trained Language Models}
\paragraph{BERT} Bidirectional Encoder Representations for Transformers (BERT) is a multi-layer Transformer encoder based language model \cite{devlin2019bert}. As opposed to directional models where the input is processed from one direction to another, the transformer encoder reads its input at once; hence, BERT learns word representations in full context (both from left and from right). BERT is trained with two pre-training objectives on large-scale unlabeled text: (i) Masked Language Modelling and (ii) Next Sentence Prediction.

A number of BERT variants are available that differ in terms of (i) their architecture, e.g. BERT-base (\textit{12-layer, 110M parameters}) and BERT-large (\textit{24-layer, 340M parameters}); (ii) whether the letter casing in its input is preserved (-cased) or not (-uncased); (iii) their masking strategy, e.g. word pieces (\textit{default)} or whole words (-whole-word-masking).

\paragraph{RoBERTa} RoBERTa \cite{liu2019RoBERTa} shares the same architecture as BERT but improves upon it via introducing a number of refinements to the training procedure, such as using more data with larger batch sizes, adopting a larger vocabulary, removal of the NSP objective and dynamic masking.

\paragraph{DistilBERT} DistilBERT was introduced by \cite{sanh2019DistilBERT}. It is created by applying knowledge distillation to BERT which is a compression technique in which a small model learns to mimic the full output distribution of the target model (in this case: BERT). DistilBERT is claimed to retain 97\% of BERT performance despite being 40\% smaller and 60\% faster, as suggested by its performance on Question Answering task.

\paragraph{GPT-2} Generatively Pre-trained Transformer (GPT-2) is a unidirectional transformer based language model trained on a dataset of 40 GB of web crawling data \cite{radford2019language}. Unlike BERT, GPT-2 works like a traditional language model where each token can only attend to its previous context. GPT-2 has four variants which differ from each other in the number of layers, ranging from 12 (small) to 48 (XL).

\section{Model} \label{sec:model}

\begin{figure}[t]
    \centering
    \includegraphics[width=0.47\textwidth]{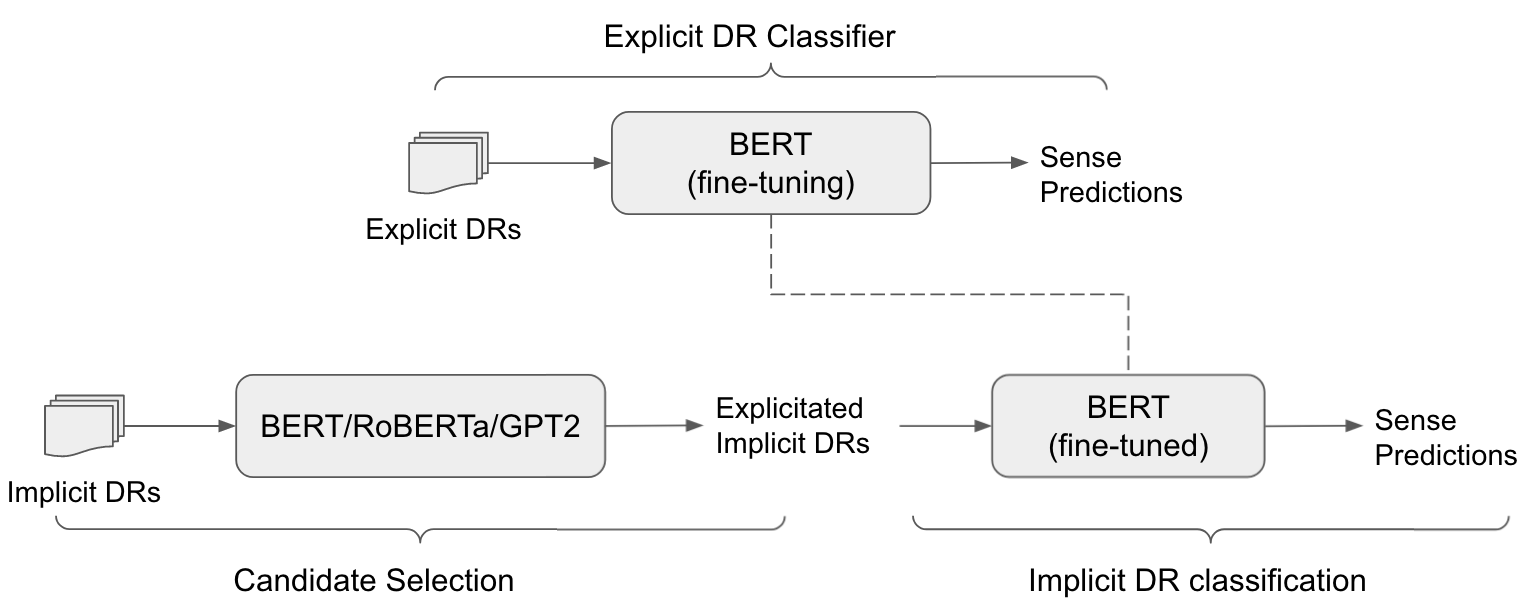}
    \caption{A high level visualization of the proposed pipeline.}
    \label{fig:pipeline}
\end{figure}

The proposed method consists of three main components: (i) a candidate generator that generates sentence pairs connected by each of a set of discourse connectives, (ii) a language model that estimates the likelihood of each candidate, and (iii) an explicit discourse relation classifier to be used on the candidates. Whole pipeline is shown in Figure \ref{fig:pipeline}. The proposed methodology does not require even a single implicit discourse relation annotation and is only distantly supervised where the supervision comes from the explicit discourse relations used in training the classifier.

The main motivation behind the proposed pipeline is the finding that discourse relations are easily classifiable if they are explicitly marked \cite{pitler2009using}. We further verify this finding via a preliminary experiment which showed that four-way classification could be performed with an F-score of 88.74 when the implicit discourse relations are \quotes{explicitated} with the \textit{gold implicit connectives} they are annotated with (see Table \ref{tab:res}). This finding is significant not only because it justifies our motivation but also shows the potential of the current approach. Secondarily, the task requires a high level understanding of the context which allows us to investigate the pre-trained language models capabilities in detecting inter-sentential relations. 

\subsection{Candidate Generation}
\label{sec:para}

Recall Example \ref{ex:1}, which contains an implicit relation between argument 1 (``\textit{A figure above \dots to expand.}'') and argument 2 (``\textit{one below \dots be ahead.}'').

Given a list of English connectives (\textit{and}, \textit{because}, \textit{but}, etc.), we generate the following explicit relation candidates for a given implicit relation:

\vspace{2mm}
\begin{tabular}{lll}
    $A_1$ & $C$ & $A_2$ \\ \hline
     \dots to expand & \textbf{and} &  [o]ne below  \dots \\
     \dots  to expand & \textbf{because} & [o]ne below  \dots \\
     \dots  to expand & \textbf{but} & [o]ne below  \dots \\
\end{tabular}
\vspace{1mm}

The list of connectives are chosen among the lexical items PDTB 2.0 annotation guideline recognizes as discourse connectives \cite{prasad2008penn}. Of the listed 100 connectives,\footnote{The modified connectives such as "partly because" are not counted as distinct types.} we limit ourselves to 65 one-word connectives to generate the candidates due to masked language models' inability to predict multiple tokens simultaneously.

\subsection{Prediction of Implicit Connectives}



Our next task is to produce a distribution over connectives $C$ conditioned on the context (arguments $A_1$ and $A_2$). For unidirectional language models (in our case: GPT-2 variants), we estimate this by computing the language model likelihood of the entire candidates and normalizing over the connectives:
\[ P_{\mathit{Conn}}(C|A_1, A_2) \propto P_{\mathit{LM}}(A_1\ C\ A_2) \]
With bidirectional masked language models (in our case: DistilBERT, BERT and RoBERTa) we need to instead provide a candidate template by inserting the special sentence separation ([SEP]) and masking ([MASK]) tokens. Then it is simply a matter of normalizing over the model's estimated probability of the connective being inserted at the position of the masking token:
\begin{multline*}
P_{\mathit{Conn}}(C|A_1, A_2) \propto\\ P_{\mathit{LM}}(C | A_1\ [\textrm{SEP}]\ [\textrm{MASK}]\ A_2\ [\textrm{SEP}])
\end{multline*}



\begin{table}[t]
\centering
\begin{tabular}{lcc}
Model&4-Way&11-Way \\ \hline \hline
PDTB train (Explicit) & 13639 & 12695\\
PDTB test & 1046& 1040\\
BioDRB(test) & 247& 140\\
BioDRB(full)  & 3001& 1755\\
\end{tabular}
\caption{Number of instances in the respective datasets. For the BioDRB test and full distinctions, please refer to Section \ref{sec:cross}.}
\label{tab:stat}
\end{table}

\subsection{Explicit Discourse Relation Classifier} \label{sec:class}

We regard discourse relation classification as a sentence pair classification task and build a classifier on top of the pre-trained BERT model from \newcite{devlin2019bert} using the recommended fine-tuning strategy. Specifically, the first and second arguments are separated via the special separator token ([SEP]) with the connective on the second argument and the [CLS] token is used for classification through a fully connected layer with softmax activation. This classifier gives us a model for the distribution $P_{\mathit{Exp}}(l|C, A_1, A_2)$ of relation labels $l$ conditioned on the connective $C$ and its arguments $A_1$ and $A_2$.

The annotation of explicit and implicit relations in the PDTB 2.0 differ in several aspects. In the case of implicit relations, PDTB 2.0 annotates arguments in the order they appear in the text, hence implicit relations can only manifest one configuration (i.e. arg1, [conn], arg2). On the other hand, the relative argument order of the explicit relations can vary to the extent that sometimes the arguments may interrupt each other (e.g. \textit{Of course}, \underline{if} \textbf{the film contained dialogue}, \textit{Mr. Lane’s Artist would be called a homeless person.} [from wsj-0039]). In order to remedy for this disparity to some extent, we only use the explicit relations which share the same relative argument order with implicit relations (i.e.\ arg1, conn, arg2) in training the classifier so that there is not any discrepancy in terms of the relation structure between training and inference phases. In total, 2558 (13.85\%) explicit relations that do not follow the (arg1,conn,arg2) order are left out.

\begin{table*}[t]
\centering
\begin{tabular}{l|cccc|c|ccc} 
Model & Temp & Cont & Comp & Exp & 4-way &\multicolumn{2}{c}{11-way} \\ \hline \hline
&\multicolumn{6}{c|}{F-score} & Acc \\ \hline
Supervised  &45.85&57.74&58.35&75.01&59.24 &39.33&55.42 \\
Gold Connective & 77.29 & 96.23 & 88.01 & 93.42 & 88.74 &57.56 &78.87 \\
Most Common Conn (but) &0.00&0.00&24.45&0.07&6.13&2.68&11.60\\
Most Common Sense            &      0.00 &         0.00 &        0.00 &      69.41 &    17.35 &  3.74  &25.89 \\  \hline

\cite{ji2015closing}& 19.26 &41.39 & 25.74& \textbf{68.08} & 38.62 & - &-\\
\cite{huang2019unsupervised}& \textbf{31.25} & \underline{48.04} & 25.15& 59.15 & 40.90 & - &-\\ 
\hline
BERT-base-uncased&14.97&29.06&32.05&59.45&33.88&13.88&23.16\\
\qquad\qquad\qquad\qquad + Margin&19.49&36.54&32.48&52.99&35.37&14.12&24.45\\
\hline
BERT-large-cased&9.33&27.06&36.89&68.58&35.47&12.29&24.55\\
\qquad\qquad\qquad\qquad + Margin&9.76&35.62&38.80&67.97&38.04&13.24&26.86\\
\hline
BERT-large-cased-wwm&10.89&36.29&42.69&62.38&38.06&16.79&27.23\\
\qquad\qquad\qquad\qquad + Margin&15.02&41.97&41.81&60.80&39.90&17.50&28.74\\
\hline
BERT-large-uncased&25.69&30.55&30.10&62.50&37.21&15.57&25.25\\
\qquad\qquad\qquad\qquad + Margin&\underline{27.32}&41.01&32.28&59.07&39.92&15.67&28.01\\
\hline
BERT-large-uncased-wwm&18.35&35.20&40.19&58.88&38.15&16.47&26.80\\
\qquad\qquad\qquad\qquad + Margin&17.27&42.93&41.16&55.61&39.24&17.26&29.17\\
\hline \hline
DistilBERT-base-cased&16.77&46.19&23.19&39.07&31.31&15.87&22.71\\
\qquad\qquad\qquad\qquad + Margin&21.09&\textbf{48.05}&29.25&37.04&33.86&16.68&26.38\\
\hline \hline
RoBERTa-base&9.65&19.64&36.57&66.72&33.14&11.67&23.54\\
\qquad\qquad\qquad\qquad + Margin&9.18&22.77&35.90&66.36&33.55&13.01&25.32\\
\hline
RoBERTa-large&10.79&30.32&\underline{48.35}&68.44&39.48&16.15&27.66\\
\qquad\qquad\qquad\qquad + Margin&13.30&33.19&\textbf{49.52}&\underline{67.90}&40.98&17.63&29.57\\
\hline \hline
GPT2&16.60&31.96&35.79&62.62&36.74&11.68&24.04\\
\qquad\qquad\qquad\qquad + Margin&18.27&37.31&35.93&61.70&38.30&13.07&26.02\\
\hline
GPT2-large&19.91&35.27&40.38&59.17&38.68&15.18&25.63\\
\qquad\qquad\qquad\qquad + Margin&23.17&40.55&40.30&60.39&\textbf{41.10}&16.03&27.50\\
\hline
GPT2-XL&21.59&30.88&40.18&63.01&38.92&16.98&26.01\\
\qquad\qquad\qquad\qquad + Margin&23.06&34.49&42.66&63.98&\underline{41.05}&18.50&28.32\\
\hline \hline
\end{tabular}
\caption{The results of the proposed methodology with various pre-trained language models. The average performance over four runs is reported (numbers within parentheses indicate the standard deviation). L stands for 'large' and wwm stands for 'whole-word-masking'. "+ Margin" refers to the second inference strategy explained in Section \ref{subsec:marginal}. Best scores are presented in bold, second bests are in italics (excluding the baselines).}
\label{tab:res}
\end{table*}

\subsection{Final Model} \label{subsec:marginal}

In our experiments we combine the models in two ways. The simplest way is a straightforward pipeline approach, where the single most likely implicit connective is predicted, and then fed to the explicit relation classifier:
\begin{multline*}
P(l|A_1, A_2) = \\ P_{\mathit{Exp}}(l|\argmax_C P_{\mathit{Conn}}(C|A_1, A_2), A_1, A_2)
\end{multline*}
Even though the level of ambiguity in English discourse connectives is relatively low, we also try to account for this ambiguity by marginalizing over all connectives:
\begin{multline*}
    P(l|A_1, A_2) = \\ \sum_C P_{\mathit{Exp}}(l|C, A_1, A_2) \times P_{\mathit{Conn}}(C|A_1, A_2)
\end{multline*}



\section{Experiments}

We follow the  experimental setting of \newcite{huang2019unsupervised} which is originally adopted by \cite{ji2015closing}. The implicit relations in the PDTB 2.0 sections 21-22 are allocated as the test set whereas the explicit relations in sections 2-20;23-24 are used as the training and 0-1 as the development set of the explicit relation classifier. The evaluation is performed for both the four first-level and the most common 11 second-level senses. For the former, we report both per-class and the macro-average F1-scores similar to \newcite{huang2019unsupervised} whereas the accuracy is also reported on the second level senses following the standard in the literature. The statistics of the used datasets are provided in Table \ref{tab:stat}.

The classifiers are implemented using the Transformers library by Huggingface \cite{wolf2019huggingface}. We use the uncased BERT large model for the explicit relation classifier (Section~\ref{sec:class}). The model is fine-tuned for ten epochs with a batch size of 16, learning rate of $5\times 10^{-6}$. To optimize the loss function, we use Adam with fixed weight decay \cite{loshchilov2018fixing} and warm-up linearly for the first 1K steps. The model is evaluated with the step size of 500 and the one with the best development performance is used as the final model. 

We mainly compare our results against the recent unsupervised studies we are aware of \cite{huang2019unsupervised,ji2015closing}. Additionally, we report the performance of a number baselines and upper bounds to put the results into a perspective:

\begin{itemize} \setlength\itemsep{0.1em}
 \item \textbf{Most Common Sense:} The performance when the most common sense of each evaluation level is predicted for every relation in the test set (Expansion for the first level; Contingency.cause for the second).

 \item \textbf{Most Common Connective:} The performance when the candidate with the most common explicit connective (\textit{but}) is selected for every relation in the test set.
 
 \item \textbf{Gold Connective:} The performance when the candidate  with the gold implicit connective is selected. This baseline also shows the upper bound of the proposed pipeline (see Section \ref{sec:model}.
 
 \item \textbf{Supervised baseline:} This is the results of the BERT classifier fine-tuned on the implicit discourse relations.
\end{itemize}

\section{Results and Discussion}

\subsection{Evaluation on PDTB}

The results are provided in Table \ref{tab:res}. Overall, the 4-way classification F-score ranges between 33.86 (DistilBERT) to 41.10 (GPT2-large) where three models outperform the previous state-of-the-art (RoBERTa-large, GPT2-large, GPT2-XL). Moreover, the performance is robust across different sense levels as suggested by its relative performance to the baselines in the more challenging 11-way classification. 

In addition to the increase in the overall performance, the most substantial gain is observed in Comparison relations where the unsupervised state-of-the-art is improved by almost 25\% points to 49.52\%, bringing it closer to the supervised baseline (58.35\%). The relatively successful performance in Comparison relations hold for all language models, suggesting that language models are good at detecting the cues for these relations.

Marginalizing over all connectives leads constant improvements with all language models. Marginalization yields average gain of 2.12\% when with BERT-variants and 2.04\% with GPT2 models. This step alters only a small portion of predictions, on average 10.1\% of the predictions change after marginalization. Relation-wise Contingency benefits from this step most with the average increase of 4.20\%. In order to have a better insight, we closely inspect the label shifts in RoBERTa-large's predictions which reveals that the most frequent label shift is from Expansion to Contingency relations (41.1\%). These changes mostly occur when there is a clear mismatch between the top connective and others following it in terms of their sense. To illustrate, Example \ref{ex:con} presents a relation, label of which was changed from \textit{Expansion} to \textit{Contingency} where the top five selected connectives were: \quotes{and},\quotes{as},\quotes{because},\quotes{since},\quotes{for}. Of these connectives, only \quotes{and} dominantly conveys Expansion whereas others commonly convey Contingency. Marginalization acts as a corrective step in such cases and saves the model from depending on the top-rank connective by allowing it to consider the connective predictions with lower ranks.
\begin{exe} 
\item \textit{Experts are predicting a big influx of new shows in 1990, when a service called "automatic number information" will become widely available.} \underline{[IMP=because]} \textbf{This service identifies each caller's phone number, and it can be used to generate instant mailing lists.} \label{ex:con}
\end{exe}

\begin{figure*}[t]{}
    \begin{subfigure}[t]{0.46\textwidth}
    \centering
\includegraphics[height=2.9in]{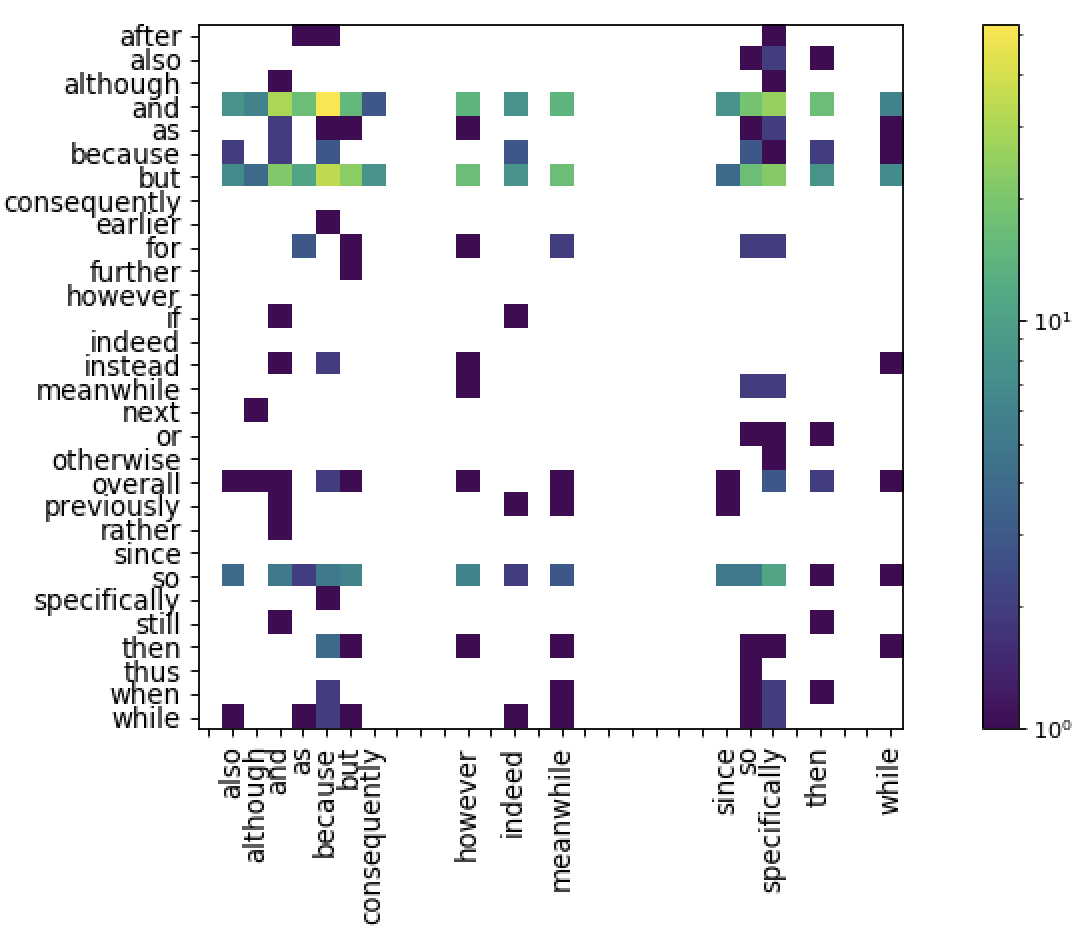}
\caption{RoBERTa-large}
    \end{subfigure}%
         ~ 
     \begin{subfigure}[t]{0.44\textwidth}
\includegraphics[height=2.9in]{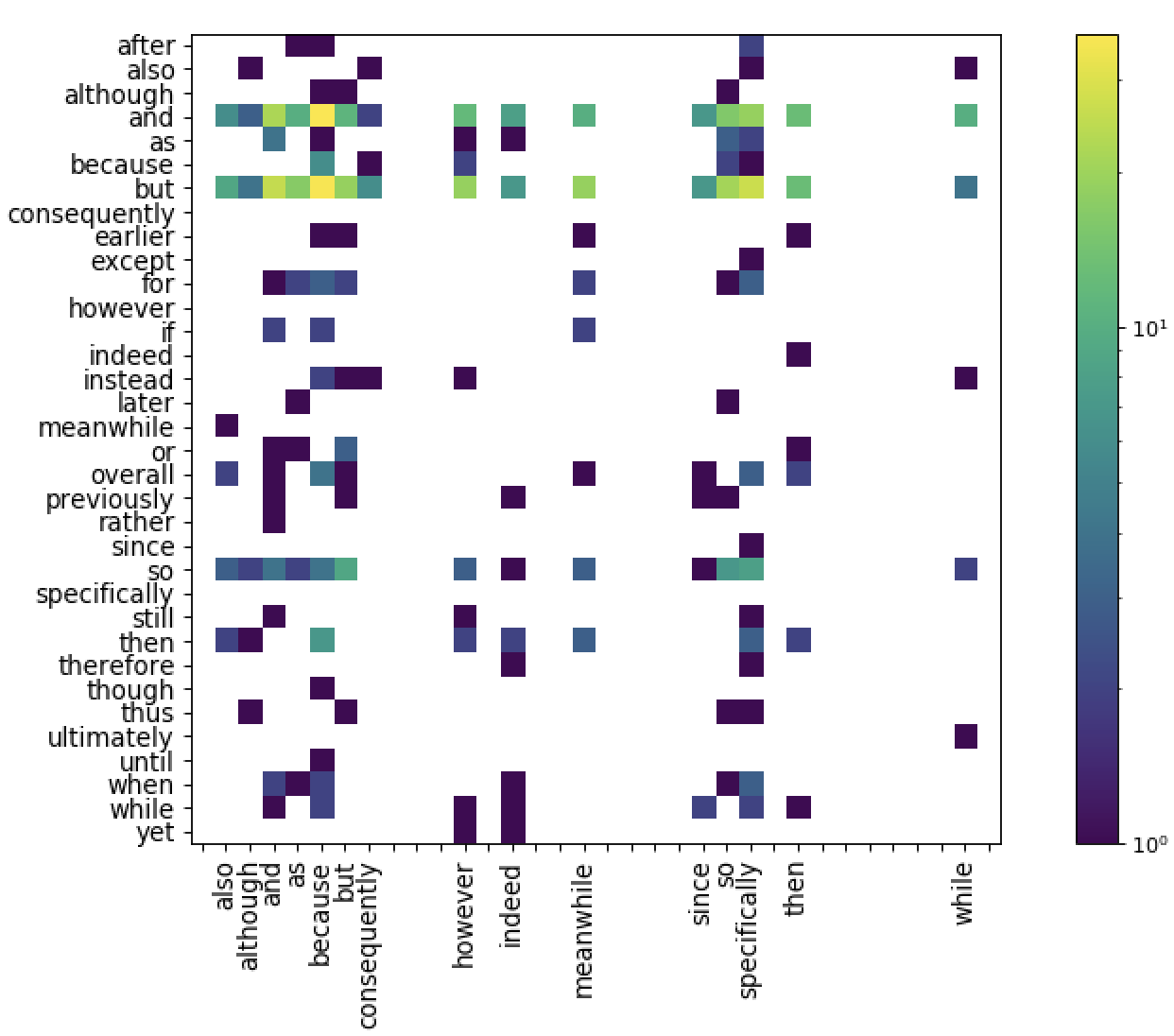}    
\caption{GPT2-large}
    \end{subfigure}%

\label{fig:results}
\caption{The (truncated) confusion matrices between the predicted and gold connectives of the implicit relations in PDTB 2.0 test set. The matrices are confined to relations with one of the most frequent 10 implicit connectives for readability purposes. The x-axis presents the gold connectives whereas the y-axis shows the predictions.}
\label{fig:cm}
\end{figure*}

Finally, as for 11-way classification, the same pattern also holds where marginalization leads to the average of 1.07\% and 2.27\% improvement in F-score and accuracy, respectively. 

\begin{table}[t]
\centering
\begin{tabular}{lcc}
Model&Conn&Sense \\ \hline \hline
always `and'& 9.38&53.15 \\
always `but'& 7.00&13.96 \\  \hline
BERT-base-cased& 14.43&38.43 \\
BERT-base-uncased& 14.85&43.40 \\
BERT-large-cased-wwm& \textbf{19.61}&\textbf{48.09} \\
BERT-large-cased& 15.69&43.31 \\
BERT-large-uncased-wwm& 16.67&45.98 \\
BERT-large-uncased& 16.39&46.85 \\ \hline
DistilBERT-base-cased& 13.45&35.18 \\ \hline
RoBERTa-base& 14.85&39.39 \\
RoBERTa-large& 17.23&46.08 \\ \hline
GPT2& 13.87&35.37 \\
GPT2-large& 14.43&39.58 \\
GPT2-XL& 14.85&39.48 \\ \hline
\end{tabular}
\caption{The agreement in percent of the language models for connective and sense prediction (see text for details). The first two rows show the results when only the respective connectives are predicted for all relations.}
\label{tab:res-con}
\end{table}

\subsection{Evaluation of the Language Models via Selected Candidates}
In order to investigate how well the language models perform their task, we present in Table \ref{tab:res-con} the agreement between the human-annotated implicit connective and each model's top-ranked connective\footnote{We limit this analysis only to the relations annotated with an one-word gold implicit connective due to our design criteria (see Section \ref{sec:para}).} (column \emph{Conn}) as well as the agreement between the most frequent sense of that top-ranked connective and the gold sense label (column \emph{Sense}). From the low connective agreement figures, we see that the models generally fail to prioritize the connective favored by the annotators; yet, as evidenced by the high sense agreement, they are able to select a connective which suits the given context and thereby helps the explicit relation classifier. We further illustrate the connective predictions of the top language models from each family (RoBERTa-large and GPT2-large) via confusion matrices in Figure \ref{fig:cm}. As can be seen, the connective predictions are very scattered showing that language models struggle to predict annotators' decisions. However, we would like to note that matching human annotators' performance in connective insertion does not yield informative insights due to ambiguity; that is, for many implicit relations, there are multiple connectives that work as fine. Therefore, we suggest the evaluation focusing on the sense conveyed by the implicit relation and the connective (column \emph{Sense}) as a more reliable way to assess the language models' performance.

too harsh a criteria to assess the language models since in many cases, there are more than one possible connectives that work as fine. Therefore, we would like to note that the second evaluation, matching the sense

Table \ref{tab:res-con} also suggests that BERT-based models perform better when it comes to selecting a suitable connective than the GPT2 family. We hypothesize that this is because bidirectional gap-filling language models have a training objective that is very close to the type of candidates we use. Finally, despite yielding the worst results, DistilBERT can retain most of BERT-base's performance ($\sim 97\%$), proving that even the smaller models can be utilized for the current task.

\begin{table*}[t]
\centering
\begin{tabular}{l|cc|cc|cc|cc} 
& \multicolumn{4}{c|}{Test set} & \multicolumn{4}{c}{Full Data} \\ \hline \hline 
& \multicolumn{2}{c|}{4-way }  & \multicolumn{2}{c|}{11-way}  & \multicolumn{2}{c|}{4-way }  & \multicolumn{2}{c}{11-way}  \\
& Acc&F1&Acc&F1&Acc&F1&Acc&F1 \\
Bi-LSTM baseline &- & -&32.97  & -&-&-&-&-\\
\cite{bai2018deep} & - & - & 29.52   & -&-&-&-&-\\
MaxEnt baseline & 58.44 & 26.64& -& -&-&-&-&-\\
\cite{shi2019next} & 77.34 & 43.03& 45.19   & -&-&-&-&-\\
\hline 
\hline 
BERT-base-uncased&54.15 & 30.29&36.98 & 14.59&54.90 & 36.30&33.80 & 13.99\\
\qquad\qquad\qquad\qquad + Margin&52.11 & 30.15&36.69 & 15.41&55.15 & 37.46&35.75 & 14.59\\
\hline
BERT-large-cased&75.37 & 26.51&37.12 & 10.29&72.28 & 30.11&32.19 & 8.28\\
\qquad\qquad\qquad\qquad + Margin&70.57 & 25.62&34.53 & 10.74&68.69 & 31.21&31.82 & 10.07\\
\hline
BERT-large-cased-wwm&62.36 & 24.59&32.95 & 10.87&65.36 & 33.59&31.81 & 11.39\\
\qquad\qquad\qquad\qquad + Margin&56.99 & 24.79&31.37 & 11.18&59.83 & 33.10&31.20 & 11.90\\
\hline
BERT-large-uncased&58.05 & 30.43&35.25 & 12.82&57.32 & 36.21&34.23 & 13.85\\
\qquad\qquad\qquad\qquad + Margin&57.24 & 31.84&37.99 & 15.58&57.01 & 37.73&35.23 & 14.54\\
\hline
BERT-large-uncased-wwm&61.22 & 32.24&\textbf{38.27} & 15.29&60.05 & 37.49&34.58 & 14.09\\
\qquad\qquad\qquad\qquad + Margin&51.95 & 30.39&36.83 & 15.62&53.98 & 37.03&34.40 & 14.55\\
\hline \hline
DistilBERT-base-cased&39.51 & 23.62&21.44 & 11.77&41.21 & 28.93&21.78 & 10.47\\
\qquad\qquad\qquad\qquad + Margin&40.00 & 27.78&25.32 & 14.97&38.35 & 30.43&23.89 & 11.56\\
\hline \hline
GPT2&59.11 & 24.36&30.94 & 11.29&62.85 & 32.37&29.82 & 10.44\\
\qquad\qquad\qquad\qquad + Margin&58.86 & 24.53&30.36 & 12.15&62.12 & 33.69&30.66 & 11.62\\
\hline
GPT2-large&62.85 & 29.70&36.69 & \textbf{19.17}&62.47 & 36.59&33.08 & 12.97\\
\qquad\qquad\qquad\qquad + Margin&60.81 & 29.48&34.82 & 15.18&61.91 & 38.33&33.86 & 13.61\\
\hline
GPT2-XL&58.86 & \textbf{33.54}&35.11 & 16.23&59.19 & 39.86&34.22 & 14.75\\
\qquad\qquad\qquad\qquad + Margin&56.75 & 33.17&34.53 & 12.54&59.19 & \textbf{41.28}&\textbf{35.33} & \textbf{15.25}\\
\hline \hline
RoBERTa-base&\textbf{78.70} & 29.70&37.84 & 12.92&\textbf{74.73}& 33.52&33.45 & 10.22\\
\qquad\qquad\qquad\qquad + Margin&78.05 & 28.83&37.41 & 13.55&74.67 & 34.31&34.13 & 10.98\\
\hline
RoBERTa-large&71.38 & 28.44&37.84 & 13.21&71.26 & 35.77&32.42 & 11.25\\
\qquad\qquad\qquad\qquad + Margin&70.98 & 28.46&38.13 & 13.49&71.42 & 37.71&33.70 & 12.93\\
\hline \hline
\end{tabular}
\caption{The results of the cross-domain experiments on BioDRB set. Test set refers to the results on the designated test set of BioDRB whereas Full data is the whole corpus. All baselines are supervised and their results are taken from \cite{shi2019next}. }
\label{tab:cross}
\end{table*}

\subsection{Cross-domain Evaluation} \label{sec:cross}

The limited number of the manual annotations does not account for the whole data bottleneck problem in discourse parsing, as the available corpora lack textual variety as much as numbers. Inarguably, PDTB is used as both the training and validation data in the bulk of studies; hence, most research on discourse parsing is confined to one domain. Unfortunately, initial attempts show that sub-tasks of discourse parsing generalize poorly across-domains \cite{stepanov2014towards}.

In order to test how our pipeline generalizes to another domain, we run a set of experiments on the Biomedical Discourse Relation Bank (BioDRB) \cite{prasad2011biomedical}. BioDRB closely follows the PDTB 2.0 annotation framework\footnote{Yet, BioDRB uses slightly different sense hierarchy. We follow the instructions on \cite{prasad2011biomedical} to map the senses back to PDTB 2.0 hierarchy.} and is annotated over 24 full-text articles in the biomedical domain which is quite different from that of PDTB. Probably due to this difference and its relatively smaller size, BioDRB is mostly overlooked in computational studies. Consequently, there are only few results on BioDRB and unsurprisingly they are all from supervised methods. We compare our results with \cite{shi2019next} which reports the state-of-the-art cross-domain results, along with the results from a number of baselines. For the sake of comparability, we follow their experimental settings and report both 4- and 11-way classification results on the BioDRB test set\footnote{which is originally suggested by \cite{xu2012connective} and consists of the files \textit{GENIA\_1421503} and \textit{GENIA\_1513057}}.

Additionally, as a more rigorous evaluation, we also report results on the whole BioDRB corpus. That way, we aim to free the evaluation of the generalization abilities of our pipeline from any bias that may rise from using a certain sub-part of the corpus. Finally, it must be noted that the LMs are not fine-tuned in any way on the target corpus (BioDRB) in either setting. The results are provided in Table \ref{tab:cross}. 


The results suggest that our pipeline has strong cross-domain performance despite explicit relation classifier's being trained on only PDTB. In both 4-way and 11-way classification, we are able to outperform the zero-shot performance of even the supervised approaches, including the recent neural approaches \cite{bai2018deep}. We hypothesize that our two-step pipeline plays the key role in mitigating the domain-specific problems. Since we are using the \quotes{raw} (unfinetuned) language models to rank candidates, we are able to directly leverage the knowledge of these models that they learn from numerous domains thanks to their diverse training data. Once the suitable connectives are highlighted by the language model, the explicit relation classifier can mainly rely on them to make the prediction; hence, less affected by the domain change.

\section{Conclusions}
In addition to its inherent difficulty, implicit discourse relation classification becomes even more challenging with the lack of sufficient data. In the current study, we focus on the latter problem by assuming the extreme low-resource scenario where there are no labeled implicit discourse relations. The data shortage is mitigated by leveraging the contextual information of the available pre-trained language models through explicitation of the implicit relations. We show that the proposed pipeline, despite its simplicity, is able to outperform the previous attempts. Furthermore, by taking another step, we tested the proposed architecture in the more challenging 11-way setting as well as on a completely different domain. The experimental results confirm that our model is robust and generalizes well, even compared to recent supervised approaches. 

\section*{Acknowledgments}
We would like to thank Dmitry Nikolaev, Johan Sjons, Bernhard Wälchli and Faruk Büyüktekin for their useful comments. The three outstanding reviews from the workshop also helped us greatly. We thank NVIDIA for their GPU grant, and the Swedish National Infrastructure For Computing (SNIC) for providing computational resources under Project 2020/33-26.


\bibliography{eacl2021}
\bibliographystyle{acl_natbib}

\end{document}